\documentclass[lettersize,journal]{IEEEtran}

\usepackage[colorlinks,linkcolor=red,anchorcolor=blue,citecolor=green,CJKbookmarks=True]{hyperref}
\usepackage{pifont}
\usepackage{booktabs}
\usepackage{graphicx}
\usepackage{float} 
\usepackage{subfig}
\usepackage{multirow}
\usepackage{soul}
\usepackage{amsmath}
\soulregister\cite7
\soulregister\ref7
\sethlcolor{white}

\begin{document}

\title{Hybrid-supervised Hypergraph-enhanced Transformer for Micro-gesture Based Emotion Recognition}

\author{
        Zhaoqiang~Xia,~\IEEEmembership{Member,~IEEE,}
        Hexiang~Huang,
        Haoyu~Chen,
        Xiaoyi~Feng,
        and~Guoying~Zhao,~\IEEEmembership{Fellow,~IEEE}
\thanks{Z. Xia is with School of Electronics and Information at Northwestern Polytechnical University, xi'an, Shaanxi 710072, China, and with Innovation Center NPU Chongqing, Chongqing 401120, China. Email: zxia@nwpu.edu.cn.}
\thanks{H. Huang and X. Feng are with School of Electronics and Information at Northwestern Polytechnical University, xi'an, Shaanxi 710072, China. Email: huanghexiang@mail.nwpu.edu.cn, fengxiao@nwpu.edu.cn.}
\thanks{H. Chen and G. Zhao are with the University of Oulu, Email: chen.haoyu@oulu.fi, guoying.zhao@oulu.fi.}
}

\markboth{arXiv}%
{Xia \MakeLowercase{\textit{et al.}}: Hybrid-supervised Hypergraph-enhanced Transformer for Micro-gesture Based Emotion Recognition}


\maketitle

\begin{abstract}
Micro-gestures are unconsciously performed body gestures that can convey the emotion states of humans and start to attract more research attention in the fields of human \hl{behavior} understanding and affective computing as an emerging topic. However, the modeling of human emotion based on micro-gestures has not been explored sufficiently. In this work, we propose to recognize the emotion states based on the micro-gestures by reconstructing the \hl{behavior} patterns with a hypergraph-enhanced Transformer in a hybrid-supervised framework. In the framework, \hl{hypergraph} Transformer based encoder and decoder are separately designed by stacking the hypergraph-enhanced self-attention and multiscale temporal convolution modules. Especially, to better capture the subtle motion of micro-gestures, we construct a decoder with additional upsampling operations for a reconstruction task in a self-supervised learning manner. We further propose a hypergraph-enhanced self-attention module where the hyperedges between skeleton joints are gradually updated to present the relationships of body joints for modeling the subtle local motion. Lastly, for exploiting the relationship between the emotion states and local motion of micro-gestures, an emotion recognition head from the output of encoder is designed with a shallow architecture and learned in a supervised way. The end-to-end framework is jointly trained in a one-stage way by comprehensively utilizing self-reconstruction and supervision information. The proposed method is evaluated on two publicly available datasets, namely iMiGUE and SMG, and achieves the best performance under multiple metrics, which is superior to the existing methods.
\end{abstract}


\IEEEpeerreviewmaketitle

\section{Introduction}
\IEEEPARstart{M}{icro-gesture} (MiG) becomes an emerging research topic in recent years \hl{as it can reveal human beings’ hidden emotional status spontaneously. They are a specific type of gestures (body movements) that are spontaneously and unconsciously elicited by inner feelings, which usually induce subtle body movements and are easily ignored by ordinary people \cite{Liu2021iMiGUEAI,Chen2023SMGAM}. The movements of MiG can be roughly categorized as self-adaptor movement such as scratching the head and rubbing the hand, defensive-behavior movement such as folding arms or moving legs, and object-related movement such as playing with clothes. Unlike ordinary body movements of the hands, head and other parts of the body that allow individuals to exhibit emotions explicitly, artistically and abstractly \cite{Karg2013Body}, micro-gestures can be used to explore the emotional states that people express intentionally and refer to some specific gestural behaviors, which are usually related to partial regions of human body and have subtle movements. So ordinary body movements are used to facilitate non-verbal communications and are helpful to be understood by others, while the MiG is an important clue to analyze the people’s hidden emotion states.} Therefore, the automatic emotion recognition with MiGs would be beneficial in the fields of social media, human-computer interaction, public safety and health care.

To analyze \hl{hidden emotions} from MiGs, the approaches of emotional artificial intelligence for \hl{ordinary body movements might be applied} to obtain the representations of local and subtle body movements. Most technologies in this \hl{field} \cite{Karg2013Body, Mahfoudi2023EmotionEI} focus on \hl{recognizing the communicative or illustrative body movements} and usually extract different feature representation from the \hl{full-body} \cite{Karg2013Body} or upper-body \cite{Noroozi2021SurveyOE} for classifying gestures into \hl{6 emotional categories, i.e., sadness, anger, happiness, disgust, fear, and surprise}. The representation for \hl{ordinary body movements} can be static or dynamic with the appearance or geometrical information. To cite a few, the geometric movement qualities \cite{Castellano2007RecognisingHE}, normalized rotation values \cite{Kleinsmith2011AutomaticRO} and Fisher score movement \cite{Samadani2014AffectiveMR} have been proposed as the representation features, and multilayer Perceptron (MLP) \cite{Kleinsmith2011AutomaticRO}, support vector machine (SVM) and hidden Markov model (HMM) \cite{Samadani2014AffectiveMR} have been used as emotion classifiers. Furthermore, various deep learning based methods like recurrent neural network (RNN) \cite{Savva2012ContinuousRO}, convolutional neural network (CNN) \cite{Kosti2017EmotionRI} and semantic auto-encoder (SAE) \cite{Wu2021GeneralizedZE} have also been introduced to analyze the temporal motion for emotions. However, these emotion-centered methods for \hl{ordinary body movements} are dedicated to the substantial movements that are intentionally performed, while the local and subtle movements of MiGs cannot be represented very well and need to be further studied by exploring the relationship between the emotion states and MiGs.

\hl{In recent years}, considerably \hl{less work has} been done on the emotion recognition with MiGs \cite{Chen2023SMGAM}. As a less explored topic, few efforts on MiGs have relatively been reported until the first dataset was published (the initial version of SMG \cite{Chen2019AnalyzeSG}). To date, only two datasets focused on different scenarios (i.e., public interview and personal interview), namely iMiGUE \cite{Liu2021iMiGUEAI} and SMG \cite{Chen2023SMGAM}, have been publicly accessed. Within the interview, the positive or negative emotions from various subjects would be disclosed by their MiGs. For the sensitive reason of biometric data, these two datasets mainly provide the skeleton data and remove the identity information in RGB videos. \hl{In the original works of two datasets \cite{Liu2021iMiGUEAI, Chen2023SMGAM}, the deep belief network (DBN) and long short-term memories (LSTMs) were separately utilized to estimate the hidden emotion states based on the micro-gestures.} More recently, based on these two datasets, some methods \cite{Li2023JointSA, Shah2023RepresentationLF, Guo2023MicrogestureOR, Yuan2023MSTCNVAEAU, Huang2023MicrogestureCB} have been developed for recognizing the categories of MiGs, rather than the emotion states. The deep models such as variational autoencoder \cite{Yuan2023MSTCNVAEAU}, the convolutional neural networks (CNNs) \cite{Li2023JointSA}, graph convolution networks (GCNs) \cite{Shah2023RepresentationLF, Guo2023MicrogestureOR} and ensemble method \cite{Huang2023MicrogestureCB} based on the skeleton data have been presented to capture the subtle movements of upper-body and \hl{classify emotions} into 17 or 33 categories. \hl{Among these methods, the GCNs based methods \cite{Shah2023RepresentationLF, Guo2023MicrogestureOR} show promising performance for the MiG analysis, which utilize the graph or hypergraph constructed from the body joints to extract the subtle motion feature.} But these methods did not further model the relationship between micro-gestures and emotion states and mainly focused on the classification of MiG categories. \hl{Some developments of employing dynamic temporal weights to the joints in a hypergraph \cite{Wei2021DynamicHC} or global systemic dynamics \cite{Zhou2024BlockGCNRT} with Euclidean distance between each joint may be helpful to improve the representation ability. However, these extension methods utilize the temporal or global motion information but ignore the local spatial relationship between each body parts (joint group, i.e., left arm and right arm), which needs to be explored with more representative modeling by considering the interaction of body joints.}

To explore the relationship between each body joints, we propose a Hybrid-supervised Hypergraph-enhanced transFormer for micrO-gesture based emotion recognition (denoted as \textbf{H2OFormer}) by reconstructing the local and subtle body movements for skeleton data. \hl{In the H2OFormer framework, the Transformer architecture is used to construct one encoder and one decoder, which are separately designed by utilizing the hypergraph-enhanced self-attention and multiscale temporal convolution for extending the vanilla hypergraph Transformer \cite{Zhou2022HypergraphTF}.} Incorporating a decoder with an encoder could introduce extra information such as the self-supervised information (i.e., the motion reconstruction), which would utilize asymmetrical blocks for the decoder compared to the encoder. \hl{Compared to the vanilla hypergraph Transformer, the hypergraph-enhanced self-attention in H2OFormer could update the relationships between hyperedges dynamically}. Based on the reconstruction, a recognition head from the output of the encoder is constructed for modeling the relationship between micro-gestures and the emotion states in a supervised way, which would exploit the shallow architecture to connect the encoded feature (local motion) and the emotion states. All these modules such as the encoder, decoder and recognition head would be cascaded and jointly learned in a one-stage strategy. The main contributions of this work are summarized as follows.
\begin{itemize}
    \item \hl{We propose a hypergraph-enhanced self-attention module by dynamically exploring the hyperedges of body joints, which can be easily embedded into the vanilla hypergraph Transformer framework for better capturing the local motion.}
    \item \hl{We design an asymmetrical encoder and decoder architecture based on hypergraph-enhanced Transformer, which accurately models the subtle movement for skeleton data based micro-gestures in a self-supervised way of motion reconstruction.}
    \item We \hl{employ a one-stage learning strategy using hybrid-supervised information to integrate the self-supervised reconstruction task with supervised recognition task}, efficiently exploring more information for sample-limited model training.
    \item We perform extensive experiments to evaluate the performance of the proposed method and achieve the best results on \hl{two} datasets, which shows its discrimination ability on micro-gesture based emotion recognition.
\end{itemize}

The rest of the paper is organized as follows. Section \ref{sec:rw} presents the related works on the emotional analysis and category classification of MiGs, \hl{and Section \ref{sec:prel} describes the preliminaries of hypergraph}. Section \ref{sec:md} introduces the deep approach in detail. Section \ref{sec:exp} provides several experiments to verify the effectiveness of the proposed deep method \hl{and discusses the characteristics of the proposed method}. Section \ref{sec:con} summarizes the entire paper \hl{and analyzes the future work briefly}.

\section{Related Work}\label{sec:rw}
Automatic emotion recognition has attracted the interest of artificial intelligence research community for the past two decades and \hl{has} developed numerous approaches by leveraging the face, body, speech or gaze of humans. \hl{Since} few works that have been discussed above focused on the emerging MiG based emotion recognition, we extend the discussion to the approaches that would potentially be used for emotion recognition based on \hl{body movements and the self-supervised learning methods for emotional or skeleton data}, which could be helpful to develop better models for recognizing MiG based emotion states. More comprehensive surveys can be revisited by \cite{Karg2013Body, Noroozi2021SurveyOE, Xu2022Emotion}. 

\subsection{Body Movement based Emotion Recognition}
\hl{According to \cite{Karg2013Body}, the body movements are usually categorized as the communicative, functional, artistic, and abstract movements, and the first type attracts most attention in past years. Except the data captured by motion and pressure sensors, the computer vision techniques due to their high availability are the focus in the context.}

As one of pioneering works, Gunes \textit{et al.} \cite{Gunes2005AffectRF} proposed to estimate the emotional categories from body movements in static frames of videos. The propositional gestures were captured by body action units tracked with Camshift algorithm and described by orientation features. By utilizing other features from faces, the emotion states were obtained by exploiting BayesNet classifiers. Then, the non-propositional movement qualities (e.g., amplitude, speed and fluidity of movement) combined with dynamic time warping based one-nearest-neighbor were used to recognize the body movements with four acted emotion states \cite{Castellano2007RecognisingHE}. With the motion capture system \cite{Venture2014RecognizingEC}, professional actors were asked to simulate emotion states and 3D motion-capture data were collected for further analyzing by representing a feature vector for each kinematics sample. Although these preliminary works can recognize the emotional gestures, they mainly \hl{aim} to explore the acted gestures and usually have huge differences with the real-world scenarios.

To achieve the non-acted (spontaneous) body postures, in \cite{Kleinsmith2011AutomaticRO}, the skeleton data in a body-movement based video game was collected by a motion capture system. Low-level movement feature description was analyzed by using normalized rotation values and then the MLP was used to recognize emotion labels defined by the authors. Similar to \cite{Kleinsmith2011AutomaticRO}, the data collected in a Nintendo sport game was used to analyze the time-related features and exploit RNNs to surpass human observers' benchmarks \cite{Savva2012ContinuousRO}. In \cite{Samadani2014AffectiveMR}, a full-body dataset was collected using a Vicon motion capture system and a stochastic model for modeling the movement dynamics was built on HMMs. A Fisher score movement representation with Hilbert-Schmidt independence criterion was then used to recognize affective \hl{movements using} SVM. Furthermore, a group of features from the full body \cite{Fourati2015MultilevelCO} were utilized to describe movements on various levels and employ random forest to predict eight emotion states in various daily actions such as walking, sitting or knocking. Senecal \textit{et al.} \cite{Senecal2016ContinuousBE} proposed to recognize continuous emotional behaviors by using Laban movement analysis mapped onto Russell circumplex model.

Apart from the professional movement-captured devices, other cameras like  Microsoft Kinect camera are also exploited to capture the movements of the whole body. By recording the gaits with Kinect \cite{Li2016EmotionRU}, features from 3D coordinates of 14 main body joints were extracted and then processed by Fourier transformation and principal component analysis. Naive Bayes, random forests and SVM were further used as classifiers to analyze various emotions. Besides, with the geometric features, Piana \textit{et al.} \cite{Piana2016AdaptiveBG} proposed to combine psychological theories, such as impulsiveness and contraction index, to implement the tasks of emotion recognition and emotion expression. Moreover, multimodal information has also been explored to discover the interrelation between face or speech and body gestures for emotion recognition. In \cite{Yang2014AnalysisOE}, prosody and audio spectral features for modeling the interaction dynamics of speech referred to three types of body representations, while the facial expression, gesture and acoustic features were used within an automatic system based on a Bayesian classifier \cite{Kessous2010MultimodalER}.

In more recent studies, deep learning based techniques begin to dominate this field. Kosti \textit{et al.} \cite{Kosti2017EmotionRI} presented to recognize emotion states based on static RGB images by using CNNs with low-rank filters. With CNNs, the person and the whole scene were jointly explored to analyze the emotion states by considering the continuous dimensions valence, rousal and dominance. In \cite{Wei2021TimeDependentBG}, an attention-based channel-wise CNN was presented to retain the independent characteristics of each body joint and learn key body gesture features in RGB videos. Besides CNNs, other deep learning models such as SAE \cite{Wu2021GeneralizedZE}, LSTM \cite{Baza2022SwarManAS}, and two-branch fusion framework \cite{Chen2023CoupledME} have also been employed to obtain good representation for sequential actions in the task of emotion recognition in videos. These deep models are borrowed from the general-purpose video understanding and directly used for the gesture based emotion recognition, \hl{which can preferably capture the motion feature by an encoder for ordinary body movements}.

\subsection{Self-supervised Learning for Skeleton Data and Emotion Recognition}
\hl{Since the encoder-decoder architecture we designed in this work refers to the self-supervised learning (SSL), we summarize several SSL methods in this section. SSL usually can learn discriminated features from various data without relying on human-annotated labels \cite{Gui2023ASO} and achieves great development in the past decade \cite{Sarkar2022SelfSupervisedER}. On one hand, SSL based on one modality, i.e., skeleton data, has been applied to various tasks like action recognition \cite{Lin2020MS2LMS, Gao2020ContrastiveSL, Men2023FocalizedCV, Guo2022ContrastiveLF, Askari2023SelfSupervisedVI, LI2023Cross}, which has the similar input of MiG based emotion recognition. On the other hand, as most datasets in the field of emotion recognition contain insufficient samples and usually limit the feature learning, SSL has also been applied to learn emotion representation for further analysis from various modalities, e.g., facial expression \cite{Roy2021SelfsupervisedCL, Shu2022RevisitingSC, Kim2022EmotionawareMC, Liu2023PosedisentangledCL, Jegorova2022SSVAERRSA, Ma2023AUA, Zhang2023MultimodalFA}, speech \cite{Goncalves2022ImprovingSE, Yang2022ExploitingFO}, gait \cite{Olugbade2024MovementRL,Lu2023SeeYE}, electrocardiograms (ECG) \cite{Sarkar2019SelfSupervisedLF,VazquezRodriguez2022TransformerBasedSL} and Electroencephalography (EEG) \cite{Zhang2021GANSERAS,Li2022GMSSGM}. As limited SSL methods have been used for skeleton data based emotion recognition, we would revisit the SSL methods for skeleton data and emotion recognition, which may be helpful to design a method for skeleton based MiG analysis.}

\hl{For utilizing the skeleton data to explore robust representations, some approaches in the task of action recognition have been presented with the SSL strategy inspired by the success of self-supervised learning in image and video tasks. As a primary work for skeleton data, MS$^2$L \cite{Lin2020MS2LMS} integrated reconstruction, classification, and projection tasks based on recurrent layers for self-learning and then obtained skeleton features from action recognition. In \cite{Gao2020ContrastiveSL}, the semantic invariance under the variant of distance and viewpoint was modeled by two separate transformations with contrastive learning of skeleton sequence to a given skeleton sequence, which maximized agreement by using the contrastive loss. Furthermore, a focalized contrastive view-invariant learning framework by gated recurrent unit based encoder and decoder \cite{Men2023FocalizedCV} was proposed to maximize the mutual information between multi-view action pairs by adapting contrastive self-learning, which leveraged a dynamic-scaled focal loss based on the geometric distance of the contrastive representations. in \cite{Guo2022ContrastiveLF},  the query, extended query and key encoders were used to extract the temporal motion features and then trained by minimizing dual distributional divergence for skeleton sequences. Besides, the skeleton data could be transformed into the image-like data and the image based SSL method can be applied directly \cite{Askari2023SelfSupervisedVI}. A cross-stream contrastive learning model \cite{LI2023Cross} was exploited by two GCN-based skeleton encoders to calculate the correspondence of intra-stream and inter-stream.} 

\begin{figure*}[ht]
    \centering
    \includegraphics[width=0.8\textwidth]{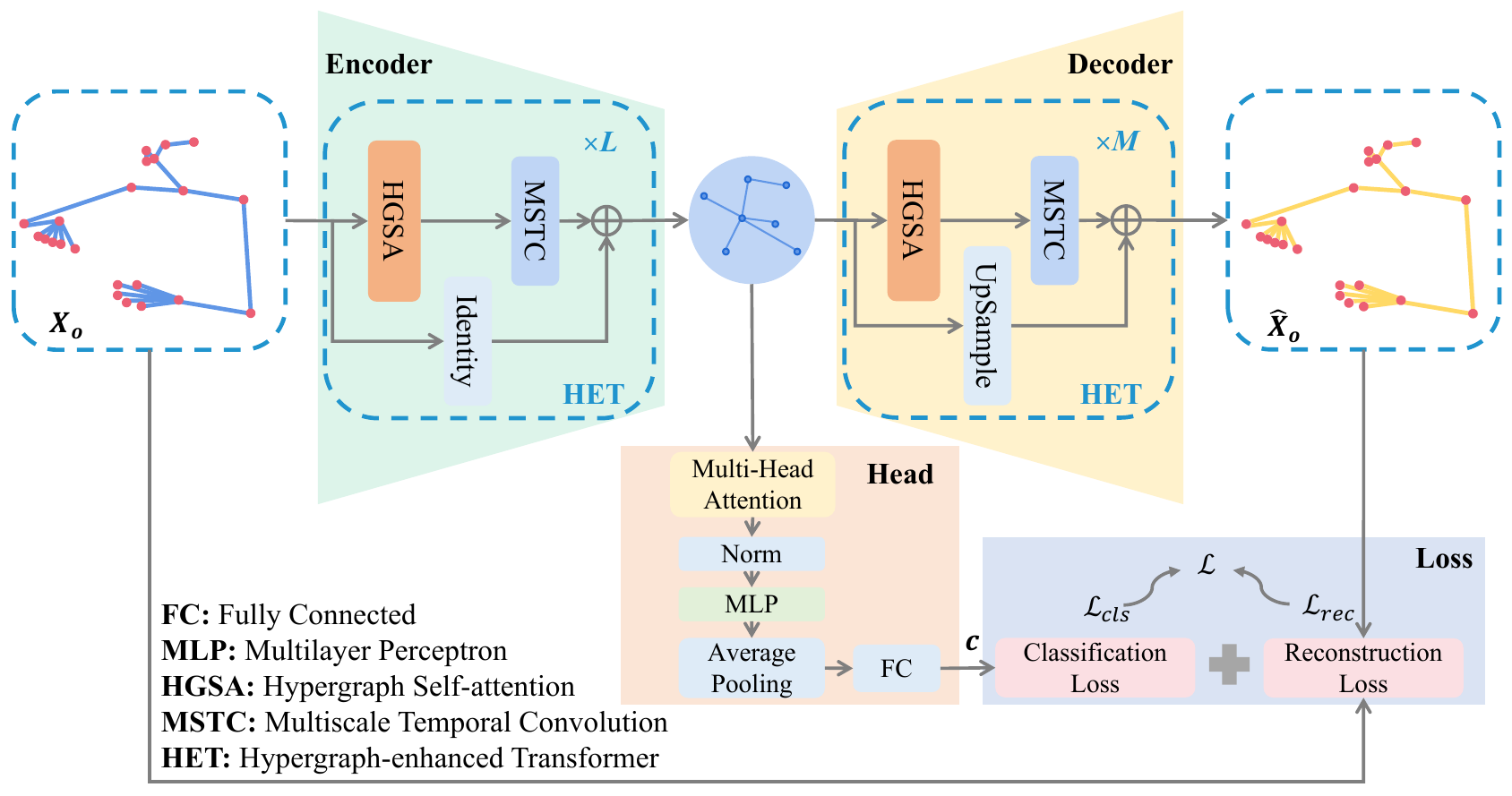}
    \caption{The architecture overview of our proposed model (H2OFormer), which is consisted of four components, i.e., the Encoder, the Decoder, the Head and the Loss.}
    \label{fig:model}
\end{figure*}

For recognizing facial expressions, the contrastive learning \cite{Roy2021SelfsupervisedCL}, a typical SSL strategy, was utilized as the first step for constructing a encoder to represent multiview facial data. Then this strategy was adequately revisited and extended into \hl{three operations, i.e., positives augmentation, hard-negatives increment, and false-negatives cancellation} \cite{Shu2022RevisitingSC}, or new feature space through feature transformation \cite{Kim2022EmotionawareMC} for obtaining expression-aware representations. To further address the issue that the contrastive learning above tends to learn pose-invariant features, a pose-disentangled contrastive learning method \cite{Liu2023PosedisentangledCL} was presented for general self-supervised facial representation. Furthermore, the contrastive SSL strategy was extended to video based emotion recognition task and one 3D convolution based encoder was constructed for capturing the temporal motion of facial regions \cite{Jegorova2022SSVAERRSA}. With another SSL strategy, i.e., masked autoencoders (MAE), MAE-Face \cite{Ma2023AUA} based on images was developed to obtain robust visual representations for facial affect analysis and can be pretrained on a large-scale face image database \cite{Zhang2023MultimodalFA}.

For recognizing emotions from other non-visual modalities, the SSL strategy was also combined with the specific architectures being adaptive to the corresponding modality. To cite a few, the SSL method \cite{Goncalves2022ImprovingSE} was employed to learn better representation for predicting emotional cues from speech with using the energy variations and facial activation as the self-supervision. Another shallow architecture with crossmodal attention fusion \cite{Yang2022ExploitingFO} was used for sentiment analysis and emotion recognition by ﬁne-tuning two pretrained self-supervised learning models. \hl{In \cite{Olugbade2024MovementRL}, two independent representation models like GCN and LSTM neural network were constructed and then used to contrastively learn pain-level representations from body movements by using a comparison network. Gait based emotion representation \cite{Lu2023SeeYE} was obtained by a cross-coordinate contrastive learning framework. In this framework, multiple encoders were learned by considering the semantic invariance with using ambiguity transform and hybrid samples memory bank.} Moreover, an ECG based emotion recognition system \cite{Sarkar2019SelfSupervisedLF} employed a signal transformation recognition network in the self-supervised learning step and then fed the transformed signal to an emotion recognition network. To exploit representations of time-series ECG signals, a Transformer based framework \cite{VazquezRodriguez2022TransformerBasedSL} by masking the ECG signals randomly as input was presented to overcome the relatively small size of datasets with emotional labels. Similar to EEG signals, to reduce the overfitting, a generative adversarial network-based self-supervised data augmentation was incorporated to combine adversarial training with self-supervised learning for EEG based emotion recognition \cite{Zhang2021GANSERAS}. To learn more general representation from multiple tasks, a graph-based multi-task self-supervised learning model \cite{Li2022GMSSGM} was built for EEG emotion recognition, which maps the transformed data into a common feature space with contrastive learning.

\hl{The above-mentioned SSL methods have been shown to improve the performance of skeleton data based action recognition and emotion recognition for 1D, 2D or 3D data. For skeleton data, most approaches construct various encoders in different views or models and perform the contrastive learning among these encoders, while few approaches try to reconstruct the actions due to the difficulty of designing decoders for skeleton sequences. For emotion tasks, most of them use the SSL strategy to obtain a pretrained model and then apply it for recognizing emotions directly in an unsupervised way (one-stage) or  fine-tune it with limited supervised information (two-stage). However, these deep models that are used with SSL are mainly focused on the significant motion from the skeleton data, which usually do not incorporate the dynamical relationships between various body joints and temporal variations for the emotional motion. So the special architecture and SSL strategy need to be further studied in the task of MiG based emotion recognition with skeleton data, which usually refers to the subtle motions for some partial joints of body.}

\section{Preliminaries}
\label{sec:prel}
\hl{In this paper, the MiG recognition framework is designed on the basis of hypergraph, which would be introduced in the following content for a start. Generally speaking, the \textit{hypergraph} is a generalization of graph in which an edge can join any number of vertices (also called as \textit{hyperedge}) for geometric-like data and has been applied in many vision-related tasks, such as data classification \cite{Gao2023HGNN} and action recognition \cite{Zhou2022HypergraphTF}.

For the human-centric tasks, by considering each joint (keypoint) of human body as a vertex $v$ and their potential connection as a hyperedge $e$ in a graph, the hypergraph $\mathcal{H}$ is constructed by containing the vertices $\mathcal{V}$ and their hyperedges $\mathcal{E}$, which is $\mathcal{H} = (\mathcal{V},\mathcal{E})$. With this representation, the spatial-temporal motion information can be extracted from this hypergraph by using multi-layer network with various architectures. The numbers of vertex and hyperedge are denoted as $|\mathcal{V}|$ and $|\mathcal{E}|$. The incidence matrix $H\in \left\{ 0,1\right\}^{|\mathcal{V}|\times|\mathcal{E}|}$ can be used to quantitatively characterize the hypergraph $\mathcal{H}$. Each entry in $H$ is denoted as:}
\begin{equation}
    H(v,e) = \left\{
    \begin{array}{rcl}
    1, if v \in e \\
    0, if v \notin e
    \end{array} \right.
\end{equation}

\hl{The people from the same scenario (i.e., from the same dataset) share the same incidence matrix. The degree of a vertex is calculated as $d(v)=\sum_e H(v,e)$, while the degree of a hyperedge is calculated as $d(e)=\sum_v H(v,e)$. The degree matrices $D_v \in \mathcal{R}^{|\mathcal{V}|\times|\mathcal{V}|}$ and $D_e \in \mathcal{R}^{|\mathcal{E}|\times|\mathcal{E}|}$ for vertices and hyperedges are then obtained by setting all the vertex degrees $d(v)$ and all the edge degrees $d(e)$ as their diagonal entries, respectively. In the initial stage of the hypergraph, one vertex only belongs to one hyperedge, which means $D_v(i)=1$, but one hyperedge may be connected to many vertices. In this case, all body joints are separately divided into $|\mathcal{E}|$ subsets.

Each vertex is represented as a feature vector and the features for the hypergraph can be denoted $X \in \mathcal{R}^{|\mathcal{V}|\times D}$. $D$ is the dimension of the vertex, which is equal to 2 or 3 if $X$ represents the original spatial coordinates of the body joints. So the subset representations for graph based neural network could be expressed as}
\begin{equation}
E = D^{-1}_eH^TXW_e
\end{equation}
\hl{where $W_e \in \mathcal{R}^{D\times|\mathcal{E}|}$ represents the learnable weight and could be assigned to each hyperedge for obtaining better representations. Based on the hypergraph, the task of emotion recognition turns to learn multiple weights for classifying the vertices on the hypergraph.}

\section{Methodology}\label{sec:md}
In this section, the deep framework of H2OFormer for MiG based emotion recognition will be introduced. The entire architecture of this framework is described firstly in Fig. \ref{fig:model} and then some core components are explained in detail in the following sections.

\subsection{Architecture Overview}
The input of the proposed method is the skeleton data detected from individuals in a review scenario, which is implemented by using the OpenPose algorithm \cite{Cao2017RealtimeM2} on RGB videos. Usually, the skeleton sequence from videos would be divided into several clips and each clip would be fed into the deep model. The skeleton data in $t$-th frame in one clip for input could be denoted as \hl{ $X^t_o=\left\{\mathbf{m}_1;\mathbf{m}_2;\dots;\mathbf{m}_K\right\}\in \mathcal{R}^{|\mathcal{V}|\times 3 }$}, where total $K$ body joints are detected by OpenPose in one frame. For each joint of body joints, the 3D spatial coordinates are recorded in a 3D vector by $\mathbf{m}_k=(x,y,z)_k$. For one sequence of MiGs with $T$ frames, the frames having $K$-joint motion could be combined together to be the input tensor \hl{$X_o=\left\{X_o^1;X_o^2;\dots;X_o^T\right\} \in \mathcal{R}^{T\times |\mathcal{V}|\times 3}$}. This tensor $X_o$ as an entire input would be fed into the H2OFormer and obtain a reconstructed sample \hl{$\hat{X_o} \in \mathcal{R}^{T\times |\mathcal{V}|\times 3 }$} as well as one emotion state $c \in \mathcal{Z}$.

The framework mainly consists of four core components, i.e., one Transformer encoder, one Transformer decoder, one recognition head, and the integrated loss function. The encoder and decoder by stacking many Transformer blocks are combined together to reconstruct the skeletal input in a self-supervised way, obtaining $\hat{X_o}$ from $X_o$. With the motion reconstruction, \hl{the local and subtle movements could be roughly captured by the encoder's representation vector for depicting MiGs.} However, the local and subtle movements may not reflect the real changes of hidden emotion states. So one emotion recognition head obtaining state $c$ is also appended into this framework and used to rectify the motion extraction from one MiG sequence by exploiting the supervised information. Different from the way that \hl{learns} the encoder first and the head secondly, we integrate them in a one-stage way and learn the deep model in one time. To learn the parameters of the deep model, the jointly integrated loss function is designed to consider the reconstruction loss and recognition loss. The encoder, decoder and the recognition head are \hl{learned} by the integrated loss in one-stage learning.

In the components of encoder and decoder, the hypergraph-enhanced Transformer module (HET) for each block is designed to construct the multilayer architecture. In each block, the HET is used as the building block and stacked directly block by block, which will be explained in detail in Section \ref{para:HET}. For the encoder, various blocks of HET can be employed to capture the local and subtle movements according to the complexity of the human skeleton data. For instance, the persons in SMG \cite{Chen2023SMGAM} have full-body movements with more complexity, which needs to choose more blocks \hl{(e.g., 10 blocks are cascaded in the encoder)}, while the persons in iMiGUE \cite{Liu2021iMiGUEAI} have upper-body movements with less complexity, which can decrease the number of blocks (e.g., 6 blocks in the encoder). For the decoder, the configuration is similar to the encoder but the motivation is very different, which is used to reconstruct the local and subtle motion. The number of blocks in decoder is quite different from the encoder, which adopts less blocks for decoder in the context (asymmetrical block configuration compared to the encoder). Apart from the HET, the upsampling operation is also necessary to the decoder for the gesture reconstruction. \hl{In this work, we choose to utilize the linear projection layer which could be inserted into the HET module easily as the upsampling layer for linearly transforming the dimensionality}. The upsampling operation could increase the dimensional size of features.

For the task of emotion recognition, only one Transformer structure similar to the vanilla architecture \cite{Vaswani2017AttentionIA} is used for prediction task as it does not need to extract much more information by considering the hypergraph, while it needs to build up the relationship between MiGs and emotion states. It becomes not very necessary to enhance the hyperedges in the prediction task and may limit the architecture complexity for head branch \cite{Guo2023MicroexpressionSW}. So the encoded feature from the encoder is utilized to predict the emotion states directly with a shallow architecture. In this context, one multi-head attention, one batch normalization, one MLP, one average pooling and one fully-connected layer are incorporated in the head to recognize the emotion states. The output of the prediction head is $c \in \left\{ 0,1\right\}$, where $1$ represents the positive state and $0$ the negative state.

\label{para:HET}
\begin{figure}[ht]
    \centering
    \includegraphics[width=\columnwidth]{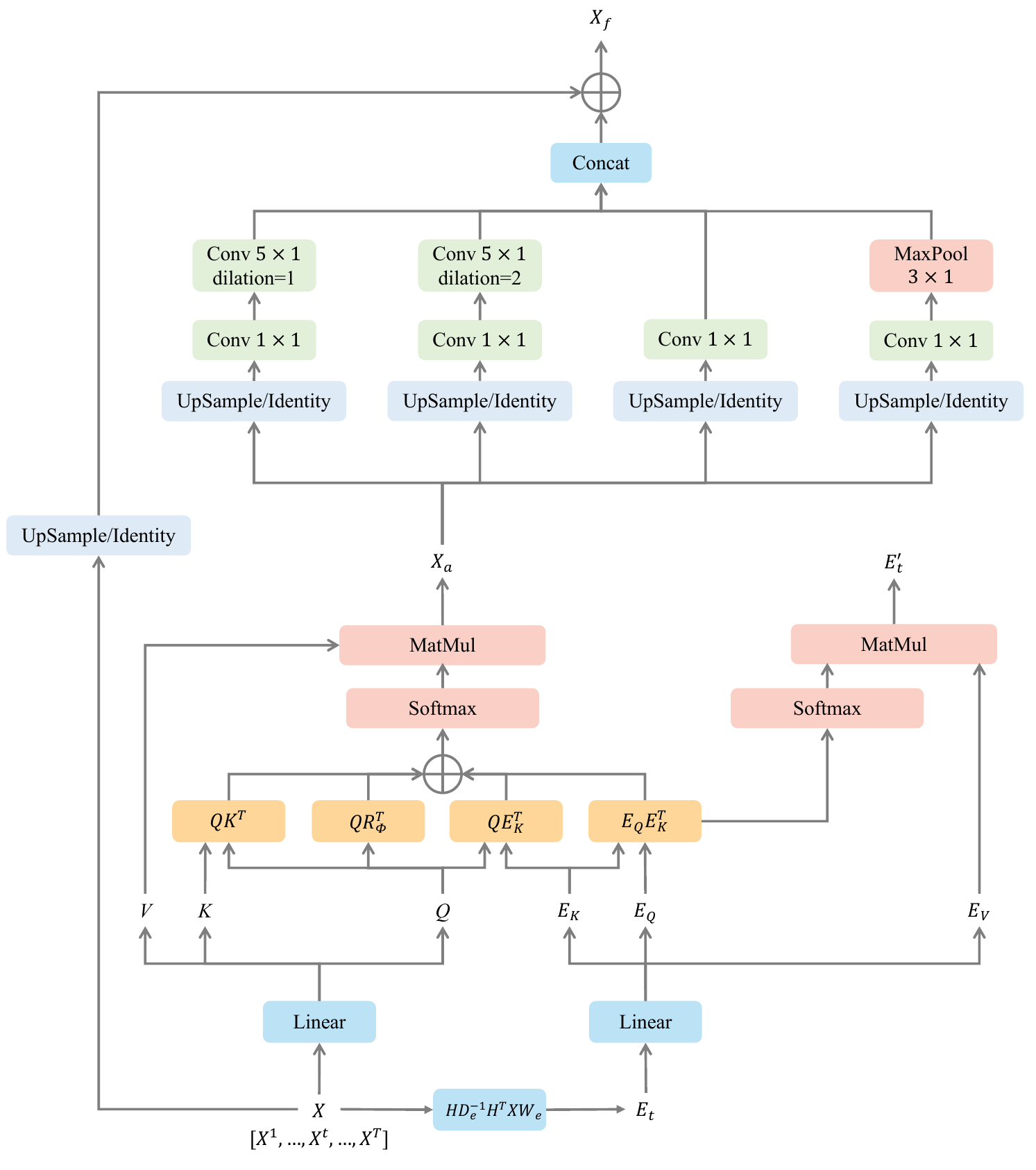}
    \caption{The hypergraph-enhanced Transformer (HET) module for each block. The upsampling operation is used only in the decoder, while the upsampling is replaced by the identity transform in the encoder.}
    \label{fig:het}
\end{figure}

\subsection{Hypergraph-enhanced Transformer Module}
The vanilla hypergraph Transformer (Hyperformer) \cite{Zhou2022HypergraphTF} is originally used for action recognition by introducing the hyperedge into the Transformer. However, the hyperedge is initially captured from the input skeleton data and fixed through the entire network learning. Whereas, as the same emotion state may contain different movements from various people, the interaction degree of body joints would be changed with various people and the connection intensity behind the body joints might be different. It means that the connection of hyperedges needs to be dynamically updated according to the different skeletal input of various MiGs. Motivated by this, we design a dynamic strategy for updating hyperedges, which is shown in Fig. \ref{fig:het}.

\hl{\textbf{Hyperedge Representation.}} Based on the incidence matrix $H$ and the original feature $X^t$ (for the input of our H2OFormer, $X^t=X_o^t$), the augmented feature representation $E_t$ for hyperedges can be further computed as:
\begin{equation}
    E_t = HD^{-1}_eH^TX^tW_e
\end{equation}
\hl{$X^t \in \mathcal{R}^{|\mathcal{V}|\times D}$} represents the $t$-th frame of input feature tensor \hl{$X \in \mathcal{R}^{T\times |\mathcal{V}|\times D }$} for each block (the output from previous block), which is equal to $X_o^t$ at first block or $X_f^t$ at $f$-th intermediate block. $D$ represents the feature dimension of \hl{each joint (keypoint)} in the hypergraph. The product of incidence matrix $H$ and the feature of joints $X^t$ can obtain the basic representation feature of hyperedges. The incidence matrix and degree matrix of hyperedges are used to normalize the weight and dimension of the hyperedges according to the original hypergraph. The learnable matrix $W_e$ could improve the representation ability by changing the dimension of input feature. The final matrix $E_t$ represents the joints with the hyperedge representation for the $t$-th frame in a MiG sequence.

\textbf{Self-attention Computation.} Based on the input $X^t$ and hyperedge representation $E_t$, the self-attention for Transformer encoder is then computed to improve the representation ability of input feature for this block. Slightly different from the hypergraph attention calculated in \cite{Zhou2022HypergraphTF}, we extend the attention computation by containing four parts as follows:
\begin{equation}
\label{eq:atten}
A=\underbrace{QR_{\phi}^T}_a+\underbrace{QK^T}_b+\underbrace{QE_K^T}_c+\underbrace{E_QE_K^T}_d
\end{equation}
The $Q$, $K$ and $V$ are obtained by linear projection from the input representation $X^t$. Similarly, $E_Q$, $E_K$ and $E_V$ with various entities are calculated by linear projection from the same hyperedge representation $E^t$, which can represent different connection weights of hyperedges for body joints. The \hl{part $a$ represents the sequential information of human skeleton. $R_{\phi}$ in part $a$ is the k-Hop relative positional embedding for sequential data, which is indexed from a learnable parameter table by the shortest path distance between the $i^{th}$ and $j^{th}$ joints \cite{Zhou2022HypergraphTF}.} The part $b$ represents the \textit{joint-to-joint} attention, which is same to the original Transformer. The part $c$ represents the \textit{joint-to-hyperedge} attention, which is induced by hypergraph attention \cite{Zhou2022HypergraphTF}. The part $d$ is newly added by our proposed method for implicitly measuring the \textit{hyperedge-to-hyperedge} attention. In each block, multihead structure is also used for the attention module. So the spatial motion information can be enhanced by using the attention feature.

\hl{In part $a$, the relative position encoding matrix $R_{\phi}\in\mathcal{R}^{|\mathcal{V}|\times |\mathcal{V}|\times D}$ of $|\mathcal{V}|$ key-points is calculated from $H_{ops}\in\mathcal{R}^{|\mathcal{V}|\times |\mathcal{V}|}$, which is given by}
\begin{equation}
\begin{aligned}
H_{ops}&=\sum_{v = 1}^{|\mathcal{V}| - 1}v\cdot(h_{v}-h_{v - 1}) \\
h_{n}&=\left\{
    \begin{array}{ll}
    I_{|\mathcal{V}|}, & if v = 0 \\
    (A_G+I_{|\mathcal{V}|})^{T}, & if 0 < v < |\mathcal{V}|
    \end{array} \right.
\end{aligned}
\end{equation}
\hl{Here, $I_{|\mathcal{V}|}$ is the identity matrix of order $|\mathcal{V}|$, and $A_G$ is the adjacency matrix of the graph. Define a matrix $R =[\vec{r}_{1},\vec{r}_{2},\cdots,\vec{r}_{m}]^{T}$, where $m$ is the maximum value of the $H_{ops}$ matrix plus one, i.e., $m=\max(H_{ops})+1$, the vector $\vec{r}_{m}\in\mathcal{R}^{D}$, and $R\in \mathcal{R}^{m\times D}$.
The relative position encoding matrix $R_{\phi}$ can be obtained as follows}
\begin{equation}
R_{\phi}=\left [\begin{array}{ccc}
\vec{r}_{hops_{00}}&\cdots&\vec{r}_{hops_{N0}}\\
\vdots&\ddots&\vdots\\
\vec{r}_{hops_{0N}}&\cdots&\vec{r}_{hops_{NN}}
\end{array}\right ]
\end{equation}
\hl{where $hops_{NN}$ is the value corresponding to the $|\mathcal{V}|$-th row and $|\mathcal{V}|$-th column in the $H_{ops}$ matrix.}

On one side, the calculated attention $A$ is further used to compute the enhanced feature $X_a^t$ by multiplying with the input as $X_a^t = Softmax(A)\cdot V$. On the other side, the hyperedge-to-hyperedge attention (part $d$) continues to be used to dynamically explore the hyperedge representation $E_t$, which can further be used as the input of next block. The updated hyperedge representation $E_t^{'}$ is calculated as
\begin{equation}
    E_t^{'} = Softmax(E_QE_K^T)\cdot E_V
\end{equation}
In \cite{Zhou2022HypergraphTF}, the hyperedge representation is only calculated in the first block and repetitively used in next blocks, while \hl{the hyperedge representation in this work could be updated in one block and passed to the next block (dynamical hyperedge representation)}. So the hyperedge representation in each block is enhanced by the hyperedge-to-hyperedge attention, which is influenced by the hyperedge with more relationships having larger attention weights.

\textbf{Multiscale Temporal Convolution.} Based on the attentive feature $X_a^t$, we further utilize the temporal convolution to extract the sequential motion information from the entire sequence $X_a=\left\{X_a^1;\dots;X_a^t;\dots;X_a^T\right\}$. As different people may have various gestures with different changing times, we employ multiscale paths in each block for multiple temporal movements. Motivated by the dilated convolution for micro-expression recognition \cite{Xia2020RevealingTI}, we use 1D dilated convolution with different sizes, i.e., $1\times 1$, $3\times 1$ and $5\times 1$. By arranging various dilated convolutions in different paths, the multiscale temporal motion information is obtained and finally concatenated into one feature presentation by adding the input feature to obtain the output feature of this block (e.g., $X_f$).

\textbf{Identity-Mapping/Upsampling.} In \hl{our hypergraph-enhanced Transformer module, multiple types of shortcut connections, i.e., identity-mapping and upsampling,} have been used from the input to the output as well as the input for temporal convolution in each block, which are shown in Fig. \ref{fig:het}. In the encoder, the goal is to extract the motion feature, so that the \hl{identity mapping connections are chosen as the shortcut connection like the ResNet, which feed the input to next layer}. In the decoder, the task is to reconstruct the local and subtle movement, so that the \hl{upsampling connections are used to increase the dimension of the input feature, which is implemented by a linear projection}. The connection type would be chosen by the block location in encoder or decoder, while they are inserted in the same position of the deep architecture for each block.

\subsection{Model Learning}
Different from the previous works, which usually used a two-stage learning strategy for self-supervised learning, we propose to learn the above-mentioned model with a one-stage learning strategy under hybrid supervision. Totally two types of tasks, i.e., reconstruction task and recognition task, are considered in the context. The reconstruction task utilizes the self-supervision to capture the subtle local motion of MiGs, while the recognition task uses the labels of emotion states to connect the MiG motion with the emotion states. Two types of losses are employed for two tasks, respectively.

For the reconstruction task, the difference between the original MiG sequence $X_o$ and the reconstructed sequence $\hat{X}_o$ is considered. For $N$ sequences with $T$ frames, the reconstruction loss $\mathcal{L}_{rec}$ is calculated as
\begin{equation}
    \mathcal{L}_{rec} = \frac{1}
    {NT}\sum_{i=1}^{N}\sum_{t=1}^{T}||\hat{X}_o^t(i)-X_o^t(i)||_2
\end{equation}
where $\hat{X}_o^t(i)$ and $X_o^t(i)$ represents the $i$-th sequence in the training set. For the recognition task, as the hidden emotion states are binary, the cross-entropy loss for emotion state classification is considered. It is calculated as
\begin{equation}
  \begin{array}{rcl}
    \mathcal{L}_{cls} = -\sum_{i=1}^N y_i\cdot log(h_{\theta}(X_o(i))) \\
    + (1 - y_i)\cdot log(1 - h_{\theta}(X_o(i))))
  \end{array}
\end{equation}
where the probability of being positive state is denoted as $p(y_i=1|X_o(i))=h_{\theta}(X_o(i))$ and the probability of being negative state $p(y_i=0|X_o(i))=1 - h_{\theta}(X_o(i))$. $h_{\theta}(\cdot)$ represents the recognition head with the encoder. To learn a model in one stage, these two losses are combined linearly by using $\mathcal{L}=\lambda_1\mathcal{L}_{rec}+\lambda_2\mathcal{L}_{cls}$, where $\lambda_1$ and $\lambda_2$ are the hyperparameters for final loss. To guarantee the one-stage learning effective, these two hyperparameters are used to ensure the same order of magnitude of two types of losses.

Besides, different from the popularly used strategy, i.e., masking, for RGB data, we choose the original skeleton data as the input of our deep model without any masking strategy by considering the structure characteristic of skeletal joints. Each joint extracted from RGB based video frames has been used to remove the redundant information for depicting the motion, which is greatly different from the RGB data. This strategy is verified in Section \ref{sec:as}. 

\section{Experiment and Discussion}\label{sec:exp}
In this section, we will introduce the datasets we used, the metrics for evaluation, the results of ablation study and comparison to the state of the art (SOTA) methods for the hidden emotion recognition with MiGs.

\subsection{Datasets and Implementation Detail}
In the experiment, we choose to utilize the iMiGUE \cite{Liu2021iMiGUEAI} and SMG \cite{Chen2023SMGAM} datasets, which are recently emerging micro-gesture datasets for hidden emotion recognition.

\begin{figure}[ht]
    \centering
    \includegraphics[width=0.9\columnwidth]{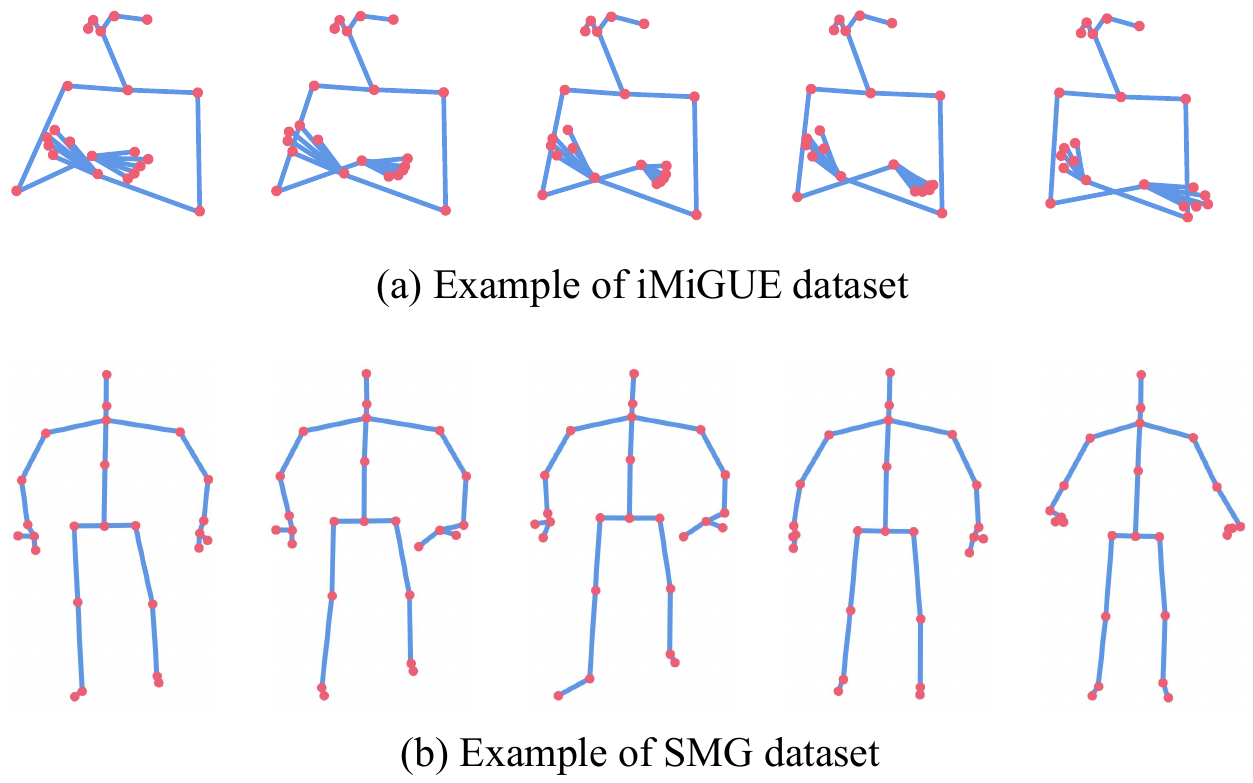}
    \caption{Examples (chosen frames from the entire sequence) of skeleton data on iMiGUE and SMG datasets.}
    \label{fig:data_example}
\end{figure}

\textbf{iMiGUE Dataset.} \hl{In this dataset, several videos containing scenarios of ``post-match press'' are collected by recording an athlete' interviews from journalists and reporters over match-related questions \& answers. The athletes have no time to prepare and should respond rapidly, which would potentially lead to positive or negative emotion states of the interviewed player as hidden emotions.} A total of 359 videos are collected from 72 subjects on online video platforms, with each video lasting approximately 6 minutes. The aggregate data from all videos consists of 3,765,600 frames, which is equivalent to approximated 35 hours of footage. The dataset incorporates skeleton data from 25 body joints, 70 facial keypoints, and 21 keypoints each for the left and right hands. The spatial coordinates of each joint or keypoint is represented by 3 numerical values. It has the average duration of 63.75 frames regarding to MiG. Additionally, there are two categories of emotion states, with 258 instances of win-status emotion (positive) and 101 instances of loss-status emotion (negative). In our following experiments, given that the majority of the data in the dataset pertains only to the upper body, we select the relevant 22 joints for the upper body from 137 available skeleton joints and keypoints. This selection includes 7 body joints, 5 facial keypoints, and 5 keypoints each for the left and right hands.

\textbf{SMG Dataset.} \hl{In this dataset, eliciting tasks by telling and repeating stories are designed to observe the MiGs when the participants need to prove that they knew the story content, respectively, no matter they are assigned with a real story or an empty story. The participants have more emotional involvement for making up a fake story and exhibit two emotional stress states as hidden emotions.} A total of 40 sequences are collected from 40 subjects with an average age of 25 years, hailing from various countries. Each sequence lasts approximated 15 minutes, and the entire dataset comprises 821,056 frames, which exceeds 8 hours of recorded data. The dataset utilizes skeleton data from 25 human body joints, \hl{with each keypoint represented by 11 numerical values, which are the 3D spatial coordinates, the 4D rotation coordinates, 2D spatial coordinates for higher-resolution images, and 2D spatial coordinates for lower-resolution images, respectively}. It has an average length of 51.3 frames for MiG. Additionally, it includes 2 categories of emotions, with 71 instances of relaxed emotion state (positive) and 71 instances of stressed emotion state (negative).

\begin{table*}[t]
\centering
	\caption{Performance comparison using various components on two datasets}
	\label{tab:components}
	\begin{tabular}{clcccccc}
	\toprule
        \multirow{2}{*}{Dataset}&\multirow{2}{*}{Method}  & \multicolumn{4}{c}{Components} & \multirow{2}{*}{Accuracy} & \multirow{2}{*}{F1-score}\\
         & &Hypergraph &Enhanced Hyperedge &Decoder Branch    &One-stage Learning & \\
        \midrule
        \multirow{6}{*}{iMiGUE} & BL &\ding{56} &\ding{56} &\ding{56}  &\ding{56}  & 0.6200 &0.5625\\
        &BL+HG &\ding{52} &\ding{56} &\ding{56}  &\ding{56}  &0.6400 &0.5682\\
        &BL+HG+EH &\ding{52} &\ding{52} &\ding{56}  &\ding{56}  &0.6400 &0.5500\\
        &BL+HG+DB &\ding{52} &\ding{56} &\ding{52}  &\ding{52}  &0.6500 &0.6337\\
        &BL+HG+EH+DB  &\ding{52} &\ding{52} &\ding{52}  &\ding{56}  &0.6700 &0.6327 \\
        &H2OFormer (Masked)  &\ding{52} &\ding{52} &\ding{52}  &\ding{52}  &0.6300 &0.6186 \\
        &\textbf{H2OFormer (Ours)}  &\ding{52} &\ding{52} &\ding{52}  &\ding{52} & \textbf{0.7000} &\textbf{0.7222}\\
        \midrule
        \multirow{6}{*}{SMG} & BL &\ding{56} &\ding{56} &\ding{56}  &\ding{56}  & 0.6316 &0.6462 \\
        &BL+HG &\ding{52} &\ding{56} &\ding{56}  &\ding{56}  & 0.6760 &0.6667\\
        &BL+HG+EH &\ding{52} &\ding{52} &\ding{56}  &\ding{56}  & 0.6842 &0.6774\\
        &BL+HG+DB &\ding{52} &\ding{56} &\ding{52}  &\ding{52}  &0.7193 &0.6957\\
        &BL+HG+EH+DB  &\ding{52} &\ding{52} &\ding{52}  &\ding{56}  & 0.6842 &0.6866\\
        &H2OFormer (Masked)  &\ding{52} &\ding{52} &\ding{52}  &\ding{52}  &0.7018 &0.7273 \\
        &\textbf{H2OFormer (Ours)}  &\ding{52} &\ding{52} &\ding{52}  &\ding{52} &\textbf{0.7544} &\textbf{0.7647}\\
		\bottomrule
	\end{tabular}
\end{table*}

\textbf{Implementation.} In our work, the parameter settings are configured as follows. The training epochs, batch size, initial learning rate and learning rate decay rate for model learning are set to 100, 64, 0.0005, and 0.1, respectively. The number of stacked encoder blocks $L$ and the number of stacked decoder blocks $M$ are asymmetrically configured to different values according the different data complexity from two datasets, i.e., $L=6$, $M=2$ for iMiGUE, and $L=10$, $M=4$ for SMG. This would be further discussed with other configurations in the ablation-study experiment for observing the influences of encoder-decoder blocks. The dimension size $D$ for the intermediate blocks is set to 216, and the temporal length $T$ for each clip is set to 52. In the self-attention module, 9 heads are used for each block. \hl{The stochastic gradient descent (SGD) is used as the optimizer for training our H2OFormer. The weights and biases of the convolution and other learnable parameters designed in H2OFormer are initialized to ensure that the gradients during the training process do not vanish or explode too quickly. Each skeleton sequence is fed into the H2OFormer and is then performed with the hypergraph convolution layer by layer. The constructed sequence and emotion category are obtained from the network and input into the loss functions for computing the gradients and updating the learnable parameters.} All experiments are performed on one NVIDIA GeForce RTX 4090. \hl{The code of the implementation is available on Github (https://github.com/xiazhaoqiang/H2OFormer-MicroGestureRec).}

\begin{figure*}[ht]
    \centering
    \includegraphics[width=0.82\linewidth]{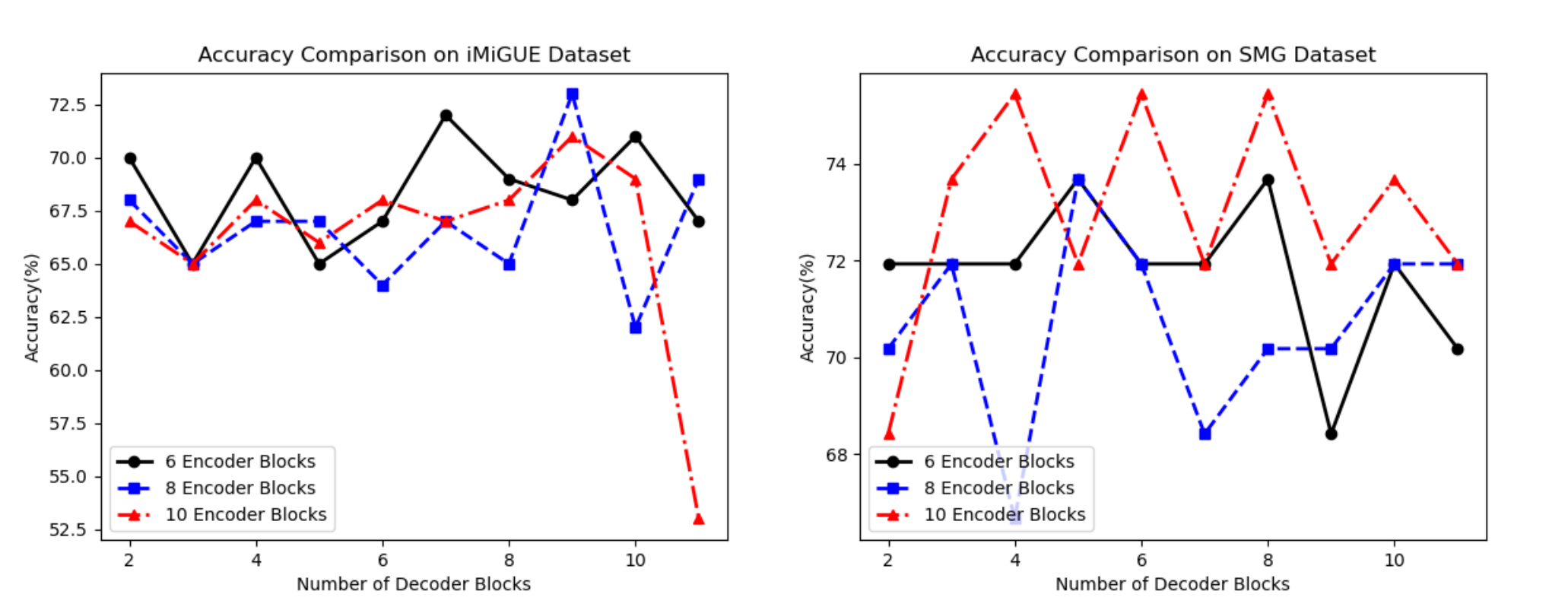}
    \caption{Performance comparison of different encoder blocks on iMiGUE and SMG dataset.}
    \label{fig:line_compare}
\end{figure*}

\begin{figure*}[ht]
    \centering
    \subfloat[]{\includegraphics[width=0.35\linewidth]{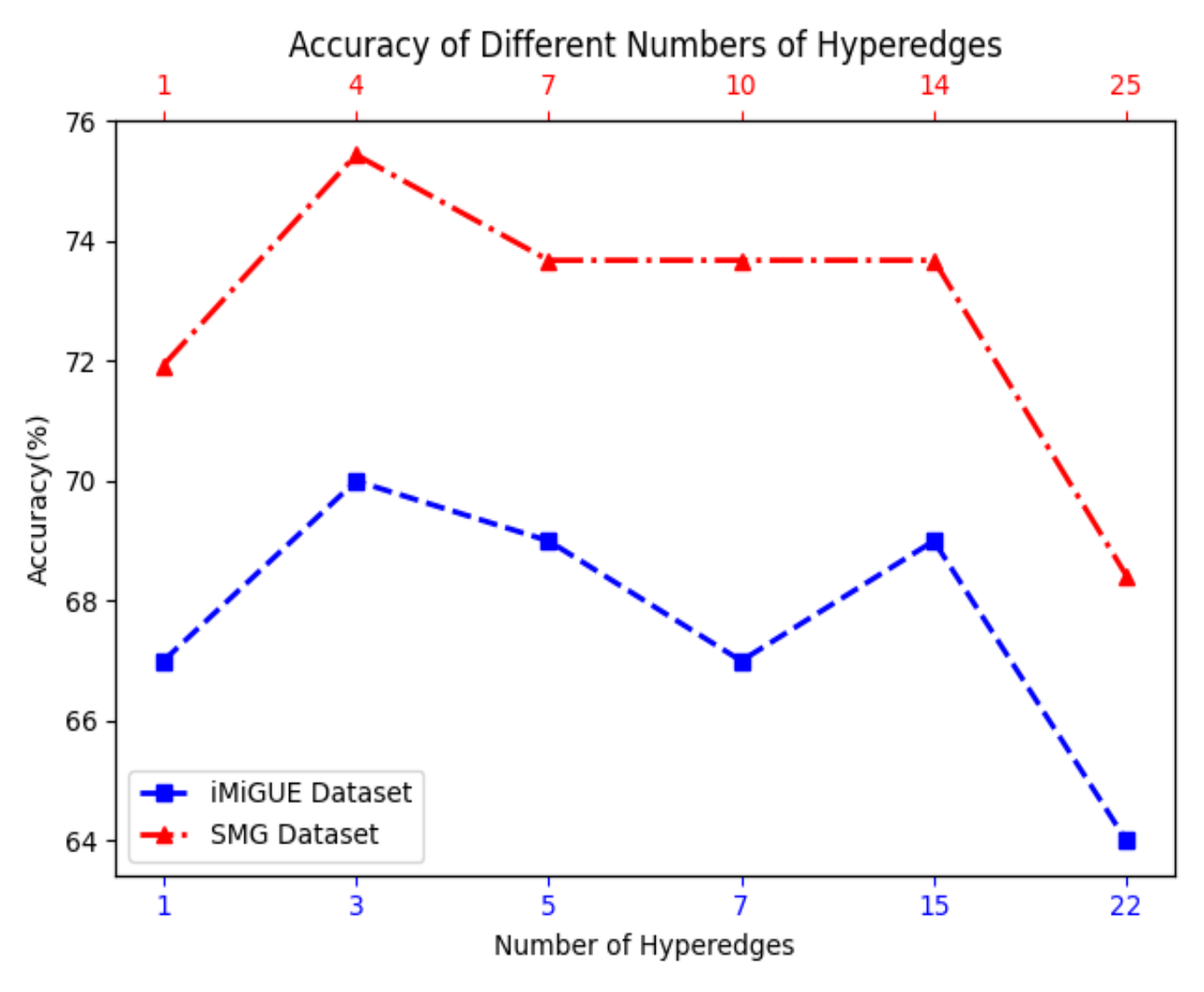}}
    \subfloat[]{\includegraphics[width=0.35\linewidth]{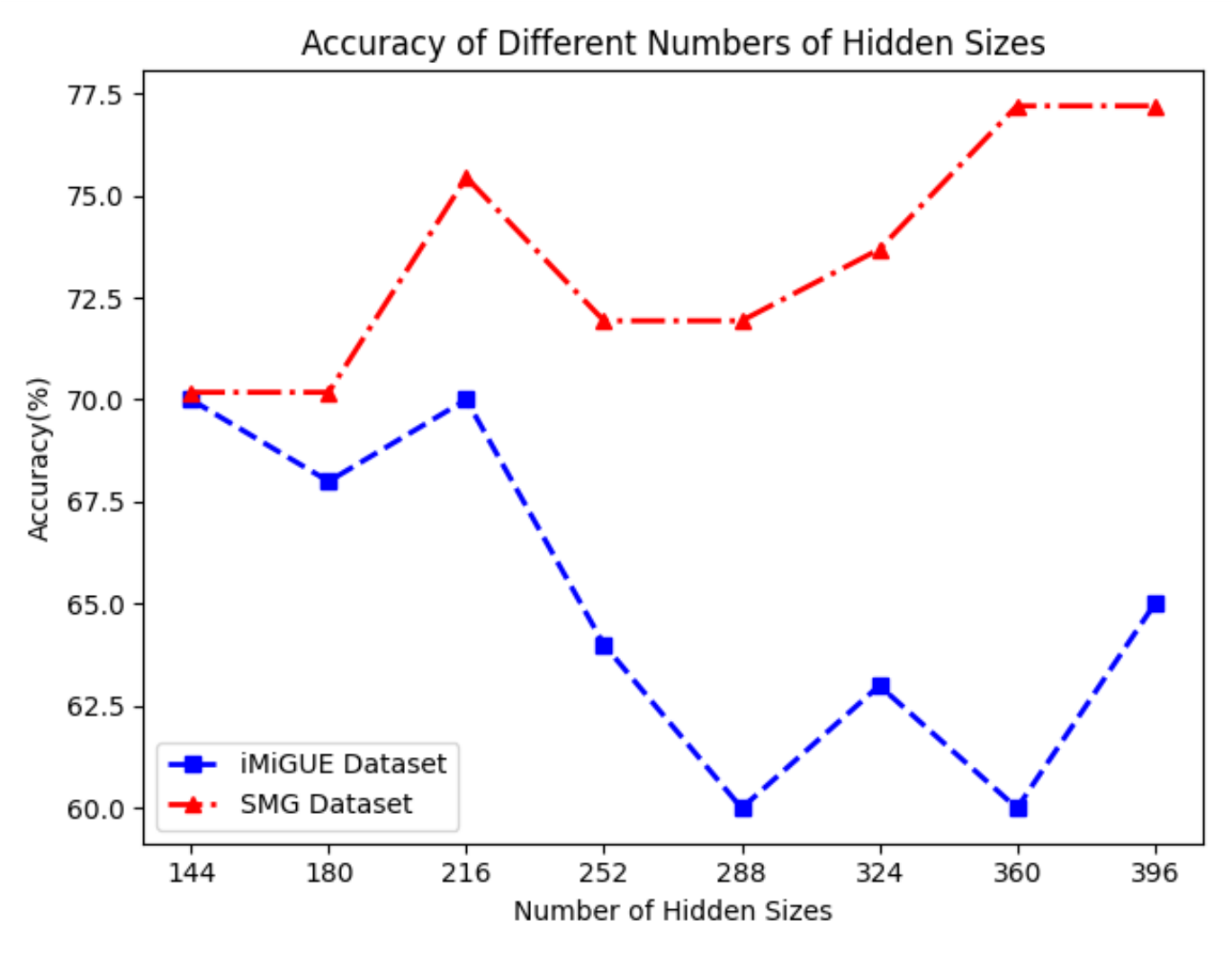}} 
    \quad 
    \subfloat[]{\includegraphics[width=0.35\linewidth]{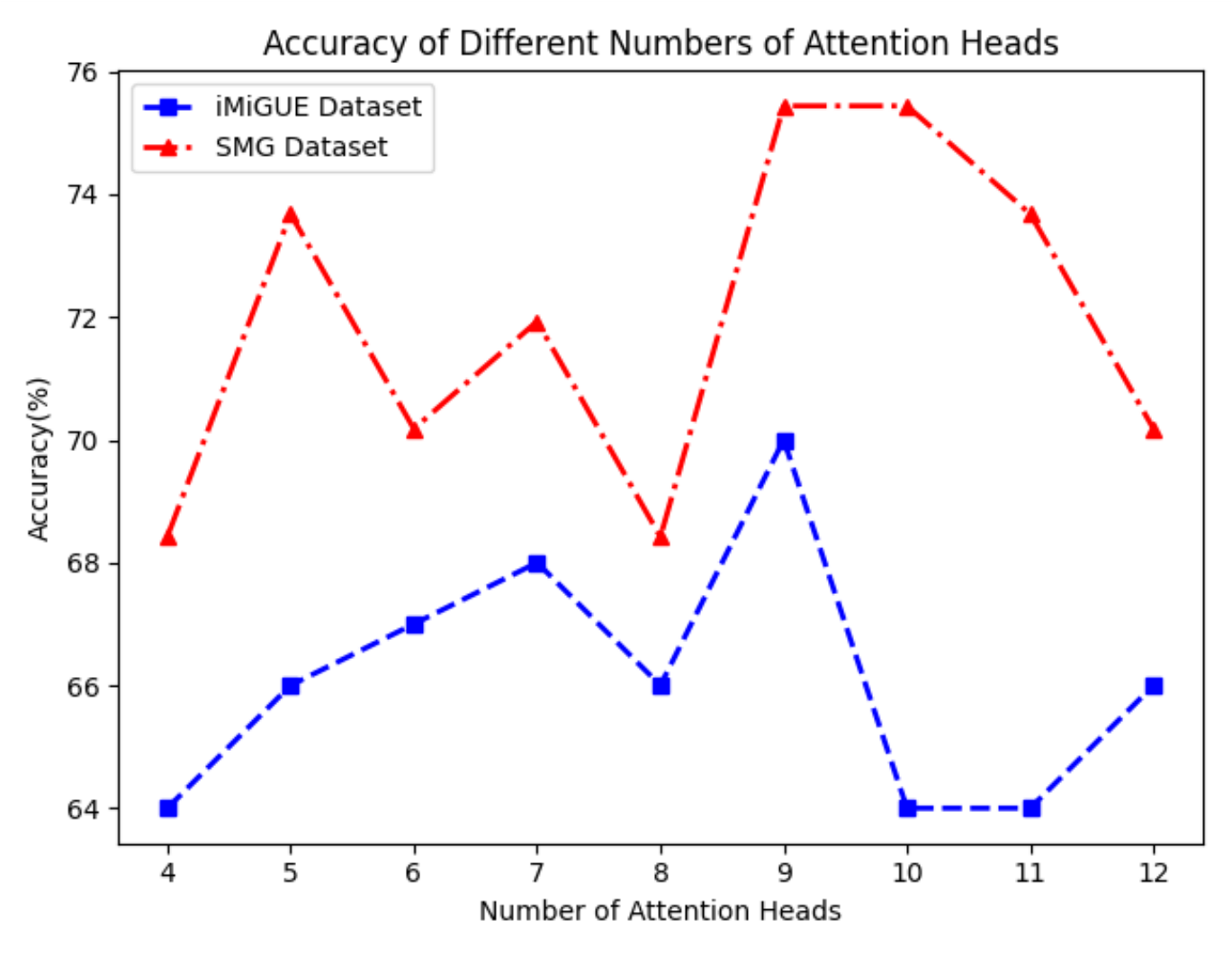}}
    \subfloat[]{\includegraphics[width=0.35\linewidth]{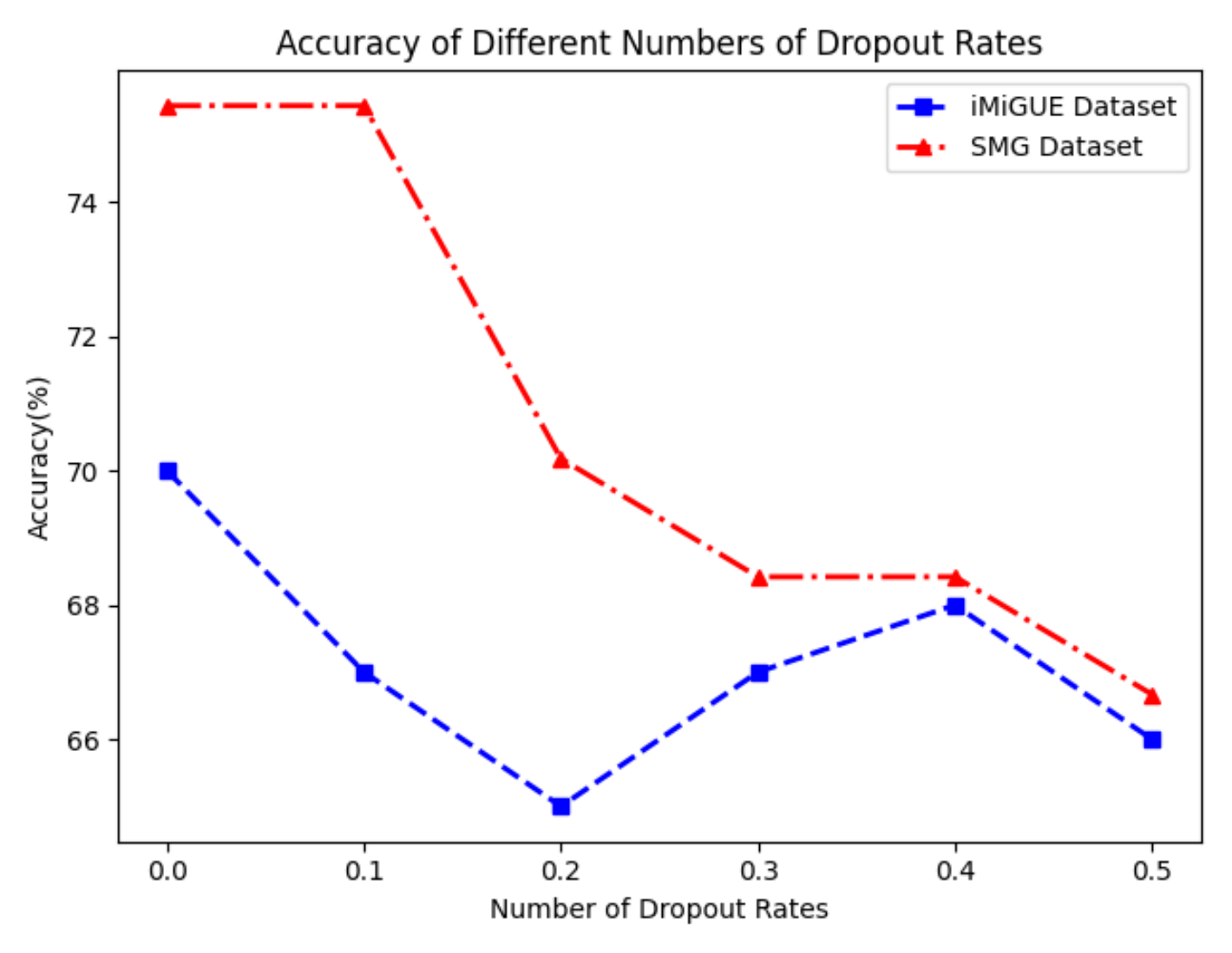}}
    \caption{\hl{Performance of using different numbers of hyperedges (a), attention heads (b), dropout rates (c) and hidden sizes (d) on iMiGUE and SMG dataset.}}
    \label{fig:compare_HG}
\end{figure*}

\textbf{Protocol.} Following \cite{Liu2021iMiGUEAI,Chen2023SMGAM}, the training set and test set are randomly chosen from all subjects, and \hl{the subject-independent protocol is used to evaluate the proposed method, which uses the data from the separate subjects (participants) for the training and test sets}. The division between training and test for subjects are kept consistent with the original datasets, which can be found in \cite{Liu2021iMiGUEAI,Chen2023SMGAM}. Within the protocol, the metrics of accuracy and F1-score are further employed to evaluate the performance of the proposed method. The accuracy is calculated as $acc = \frac{1}{N}\sum_{c} TP_c$ where $TP_c$ and $N$ are the number of true positives for $c$-th emotion state and all testing samples. The $F1-score$ is computed as $F1= \frac{2P\times R}{P+R}$, where $P = \frac{\sum_{c}TP_c}{\sum_{c}(TP_c+FP_c)}$ and $R = \frac{\sum_{c} TP_c}{\sum_{c}(TP_c+FN_c)}$ for $c$-th emotion state. $FP_c$ and $FN_c$ denote the false positives and false negatives in $c$-th emotion state.

\subsection{Ablation Study}
\label{sec:as}
Here, we choose to evaluate the important components of our proposed framework, e.g., the hypergraph used, the enhanced hyperedge, the decoder for reconstruction, and the way of training. Furthermore, the influences of various parameter configurations, e.g., the blocks of encoder and decoder and the number of hyperedges are also verified. The visualization for the hyperedge representation is also shown in this section.

\textbf{Various Components of H2OFormer.} In order to verify the effectiveness of important components in our proposed model, we conduct a series of ablation study experiments. The vanilla Transformer without using any hypergraph and reconstruction task are used as the baseline (denoted as \textbf{BL}). Four core components are sequentially added into the BL method. The component of using hypergraph (HG) is evaluated by adding the HG with BL (denoted as \textbf{BL+HG}). The component of using enhanced hyperedge with HG is verified by adding the enhanced module with BL+HG (denoted as \textbf{BL+HG+EH}) while the component of merely using decoder branch is denoted as \textbf{BL+HG+DB}, which would be trained with one-stage learning strategy. To observe the effectiveness of using one-stage learning strategy, we combine the EH and DB together (denoted as \textbf{BL+HG+EH+DB}) in a two-stage training strategy, which trains the encoder-decoder model firstly and then fine-tunes the encoder and the recognition head with the loss $\mathcal{L}_{rec}$ and $\mathcal{L}_{cls}$, respectively. \hl{The two-stage training is performed on the same dataset without any additional data.} Besides, as the masking strategy has been used frequently for RGB images \cite{He2022Masked}, we also investigate the masking strategy on skeleton data, which is denoted as \textbf{H2OFormer (Masked)}. \hl{In this method, the spatial coordinates of several joints are randomly chosen and set to zeros, in which $30\%$ joints have been masked following \cite{He2022Masked}}. The decoder will refill these missing joints with our proposed one-stage H2OFormer. Our method utilizes all the components above and has the same configuration with these baseline methods. All the ablation study methods are performed on two datasets with two kinds of metrics, i.e., accuracy and F1-score.

The experimental results of using various components are summarized in Table \ref{tab:components}. Take iMiGUE dataset for example. From the experiments of each component, it can be seen that the performance of using hypergraph can be improved as the hyperedges could focus on the local movements. The accuracy of the model can further be improved by 2\% when the decoder is added to reconstruct the data, while it is also improved by 1\% when the enhanced hyperedge is incorporated. It illustrates that the component of enhanced hyperedge could promote the ability of representation and the component of decoder could make the model focusing on the characteristics of MiG motion. It is worth noting that by jointly using the decoder of reconstructed task with the iterative hyperedge enhancement and one-stage training, the performance of the basic model is greatly improved by 8\%, much better than using them individually. Another thing that the masking strategy does not work very well on skeleton data needs to be focused. That might because the skeleton data has more compact representation as each joint will be usefully for recognizing emotion states while some regions in RGB data may be removed due to the information redundancy. The similar conclusion can be observed on SMG datasets. Only one difference on SMG dataset is that using decoder module seems more important than other modules.

\textbf{Various Blocks of H2OFormer.} To observe the influences of using different blocks for encoder and decoder, some experiments are performed on two datasets under the metric of accuracy, which is shown in Fig. \ref{fig:line_compare}. In this figure, the encoder and decoder with various blocks are employed. On these two datasets, the numbers of blocks are selected as 6, 8 and 10, respectively, with the decoders starting from 2 to 10 blocks. From the results, different patterns can be observed on two datasets. On iMiGUE dataset, the matching between the encoder and the decoder becomes important. With different blocks in encoder and decoder, the accuracy changes obviously. The models with larger complexity on iMiGUE may not achieve obvious performance improvement by adding blocks in encoder and decoder directly as the skeleton data over this dataset only contain upper-body joints. By considering the performance and the complexity of the proposed model, 6 blocks for encoder and 2 blocks for decoder are chosen as the final structure for other experiments, even if this is not the best result reported in the figure. On SMG dataset, the performance of the proposed models having various blocks changes differently. \hl{As the subjects in SMG dataset have full-body joints (more keypoints) than the iMiGUE dataset, the model having more blocks may be better than the simpler one (i.e., the model on iMiGUE dataset).} But it does mean that the more the blocks, the better the performance. The encoders with 10 blocks is matched with the decoder with 4 blocks, which also considers the performance and the complexity. \hl{However, these results also reflect that the generalization capability of H2OFormer is also slightly affected by various blocks. It would be better to know the data complexity before choosing the optimal number of blocks for our proposed H2OFormer.}

\textbf{Various Hyperedges of H2OFormer.} The hyperedges in our proposed method have important impact on the representation ability of deep model, so we investigate the number of hyperedges on two datasets, which is shown in Fig. \ref{fig:compare_HG} (a). From the figure, we can see that using too less or too many hyperedges cannot achieve good performance. \hl{On one hand, choosing too less hyperedges means that one joint or keypoint can only be connected to limited joints or keypoints. Take the movements from iMiGUE dataset for example. If choosing too less hyperedges, the keypoints on left arm and hand could not be connected to the keypoints of right arm and hand, which cannot model the joint motion of left and right arms effectively. In other words, the left arm and hand may move with the right arm and hand simultaneously for the upper body. So the joints in left arm and hand could be connected with the joints in right arm and hand. On the other hand, choosing too many hyperedges means that one joint or keypoint can be connected to almost arbitrary joints or keypoints. For example, the keypoints on left arm and hand could be connected to the keypoints of abdomen, which has a large redundancy for the model learning. Slightly different from iMiGUE dataset, the configuration for the connection of body joints could be changed on the SMG dataset, as the SMG collects the full-body data. Overall, the best configuration of hyperedges seems to be consistent with the prior structure of human body.}

\hl{\textbf{Other Hyperparameters of H2OFormer.} Here, we further evaluate the impact of using various hidden sizes, different number of attention heads for the model and the usage of dropout learning strategy for model learning. As shown in Fig. \ref{fig:compare_HG} (b), various hidden sizes can affect the recognition performance. Observed from the results on two datasets, when the hidden size is relatively small, the performance on the iMiGUE dataset is better. As the  hidden size increases, the accuracy continuously decreases. On the contrary, on the SMG dataset, as the hidden size increases, the accuracy continuously increases. For the Fig. \ref{fig:compare_HG} (c), different number of attention heads can affect the recognition performance. Observed from the results on two datasets, 9 heads used in our implementation part are the best choice for learning a good model. From the Fig. \ref{fig:compare_HG} (d), in which the dropout rate ranges from 0.0 to 0.5, it can be concluded that the dropout strategy is not suitable for the MiGs based emotion recognition. The reason may be that the body joints are connected to each other and densely modeled by the graph convolution, which could not be set to zero by the dropout operation.} 

\begin{figure}
    \includegraphics[width=0.95\columnwidth]{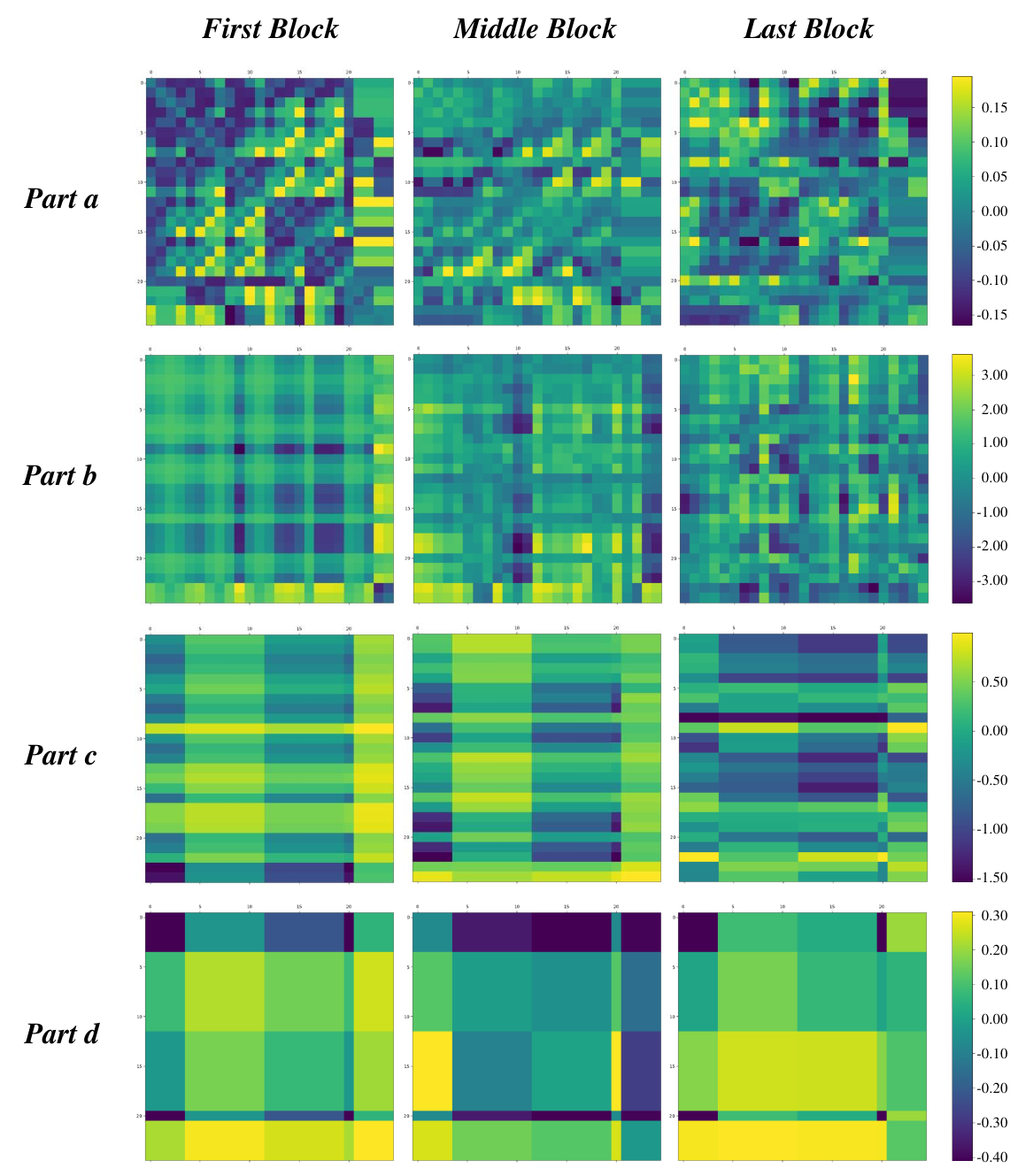}
    \caption{\hl{The visualization of four attention parts $a$, $b$, $c$ and $d$ within the encoder}.}
    \label{fig:visual_attn}
\end{figure}

\begin{figure}
    \includegraphics[width=0.95\columnwidth]{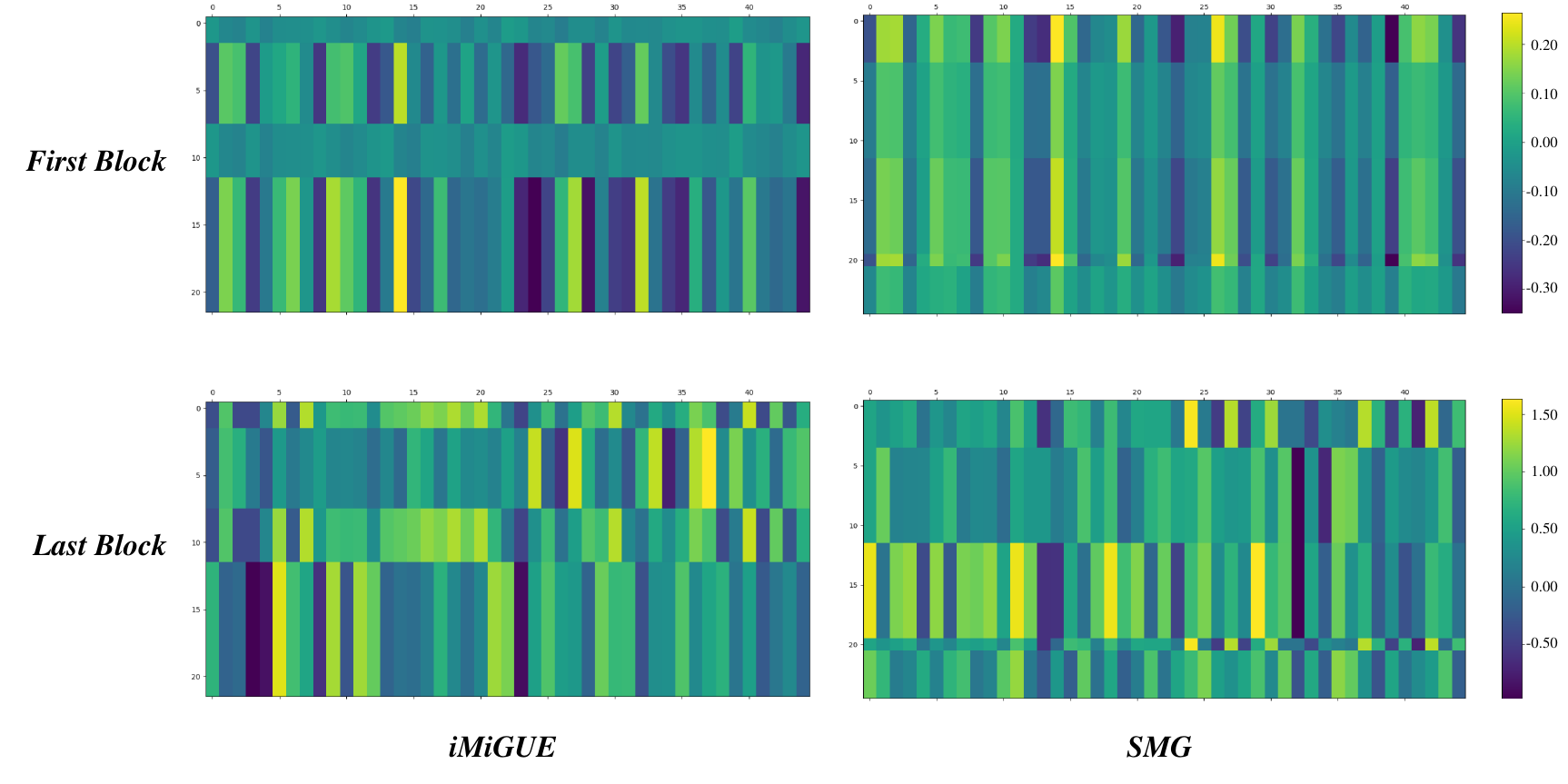}
    \caption{The visualization of hyperedge features in the first and last blocks \hl{within the encoder}.}
    \label{fig:visual_E}
\end{figure}

\textbf{The Visualization of Hyperedge Representation.} As the attention module with \hl{dynamic hyperedges} are important in the proposed H2OFormer, we illustrate out four attention parts in Eq. \ref{eq:atten}. The four parts are shown in Fig. \ref{fig:visual_attn}, which selectively shows three levels of the beginning, intermediate and final blocks in the encoder. Among four parts, we can see that different-level attention has been explored, which has progressively coarser levels of attention sequentially from part $a$ to part $d$. \hl{All the weights for the attention parts have been changed during the model learning on two datasets.} Besides, the attention weights are also dynamically changed with various blocks \hl{within the encoder}. Similar to the attention module, the hyperedges are updated dynamically and have the \hl{obvious visual changes} by comparing the first block and last block. The visual results on two datasets can be directly observed in Fig. \ref{fig:visual_E}.

\subsection{Comparison to SOTA Methods}
As currently few studies focus on the recognition of emotion states based on MiGs, we utilize the results as the baseline methods reported from two datasets \cite{Chen2019AnalyzeSG,Liu2021iMiGUEAI} for performance comparison. Except that, we also implement one of the best MiG recognition method (i.e., PoseC3D \cite{Li2023JointSA}) and the vanilla hypergraph Transformer for action recognition method (i.e., Hyperformer \cite{Zhou2022HypergraphTF}) in the task of MiG based emotion recognition. All these methods are reported with the same protocol and metric. In order to compare these methods fairly, the metric of accuracy used in \cite{Chen2019AnalyzeSG,Liu2021iMiGUEAI} is chosen in the experiment.

\begin{table}[tb]
\centering
	\caption{Performance comparison to SOTA methods on iMiGUE dataset}
	\label{tab:sota_imigue}
	\begin{tabular}{lccc}
		\toprule
        Method      &Model                      &Modality               & Accuracy\\
        \midrule
        TRN \cite{Zhou2018TRN} &\multirow{3}*{CNN}         &\multirow{3}*{RGB}     &0.53\\
        TSM \cite{Lin2019TSM}   &~               &~       &0.53\\
        I3D \cite{Carreira2017I3D} &~ &~ &0.57\\
        \midrule
        PoseC3D* \cite{Li2023JointSA}     &CNN                        &Skeleton               &0.56\\
        \midrule        
        U-SVAE+LSTM \cite{Liu2021iMiGUEAI} &\multirow{2}*{MiG+RNN}    &\multirow{2}*{Skeleton}     &0.55\\
        TSM+LSTM \cite{Liu2021iMiGUEAI}    &~                          &~                      &0.60\\
        \midrule
        ST-GCN \cite{Yan2018SpatialTG}      &\multirow{4}*{GCN}         &\multirow{4}*{Skeleton}&0.50\\
        MS-G3D \cite{Peng2019LearningGC}      &~                          &~                      &0.55\\
        Hyperformer* \cite{Zhou2022HypergraphTF}     &~                      &~                      &0.60\\
        H2OFormer (Ours)   &~                    &~                      &\textbf{0.70}\\
		\bottomrule \\
        \multicolumn{4}{l}{* represents the self-implemented method.}
	\end{tabular}
\end{table}

\begin{table}[ht]
\centering
	\caption{Performance comparison to SOTA methods on SMG dataset}
	\label{tab:sota_smg}
	\begin{tabular}{lccc}
		\toprule
        Method                              &Model                      &Modality               & Accuracy\\
        \midrule
        \multirow{2}*{Common people \cite{Chen2023SMGAM}}        &\multirow{4}*{Human}       &Skeleton                    &0.48\\
        ~                                   &~                          &RGB               &0.53\\
        \multirow{2}*{Trained evaluators \cite{Chen2023SMGAM}}   &~                          &Skeleton                    &0.66\\
        ~                                   &~                          &RGB               &\underline{0.75}\\
        \midrule
        MiG+L2GCN \cite{You2020L2GCN}        &\multirow{2}*{MiG+GCN}         &\multirow{2}*{Skeleton}     &0.47\\
        MiG+BGCN \cite{zhang2019bayesian}    &~                          &~                      &0.53\\
        \midrule
        MiG+BayesianNet \cite{Chen2019AnalyzeSG}  &\multirow{2}*{MiG+Classifier}  &\multirow{2}*{Skeleton}     &0.66\\
        MiG+WSGN \cite{Chen2023SMGAM}        &~                          &~                      &0.68\\
        \midrule
        PoseC3D* \cite{Li2023JointSA}     &CNN                        &Skeleton               &0.63\\
        \midrule
        ST-GCN \cite{Yan2018SpatialTG}       &\multirow{5}*{GCN}         &\multirow{5}*{Skeleton}&0.42\\
        MS-G3D \cite{Liu2020DisentanglingAU} &~                          &~                      &0.49\\
        NAS-GCN \cite{Peng2019LearningGC}    &~                          &~                      &0.52\\
        Hyperformer* \cite{Zhou2022HypergraphTF}    &~                          &~                      &0.64\\
        H2OFormer (Ours)                     &~                          &~                      &\textbf{0.75}\\
		\bottomrule \\
        \multicolumn{4}{l}{* represents the self-implemented method.}
	\end{tabular}
\end{table}

On iMiGUE dataset, deep models with various architectures are utilized to recognize the emotion states, which could be observed in Table \ref{tab:sota_imigue}. The \hl{CNN, RNN and GCN architectures (identified in column 2 of Table \ref{tab:sota_imigue})} are separately used to extract the feature of local motion based on skeleton data. All these methods use the spatial coordinates as the input and predict the emotion states in an end-to-end way. The macro-gesture based GCN (i.e., ST-GCN \cite{Yan2018SpatialTG} and MS-G3D \cite{Peng2019LearningGC}) and CNN (i.e., PoseC3D \cite{Li2023JointSA}) can achieve the almost similar performance, which would be surpassed by modeling the temporal motion with LSTM (i.e., U-SVAE+LSTM and TSM+LSTM \cite{Liu2021iMiGUEAI}) or Transformer (i.e., Hyperformer \cite{Zhou2022HypergraphTF}). With the enhancement of hyperedges and decoder, our proposed method can achieve better results than LSTM and Transformer based methods as the local and subtle movements can be modeled more precisely. In addition to the skeleton data, the data with RGB modality in original datasets are also used to recognize the emotion states by exploring CNN based architectures (i.e., TRN \cite{Zhou2018TRN}, TSM \cite{Lin2019TSM} and I3D \cite{Carreira2017I3D}) while the privacy issue directly using RGB data becomes severe. Besides, the performance of skeleton data based methods has been superior to the RGB data based methods. So our proposed H2OFormer outperforms all other methods on iMiGUE dataset.

On SMG dataset, the handcrafted or deep features with neural networks (i.e., BayesianNet \cite{Chen2019AnalyzeSG}, WSGN \cite{Chen2023SMGAM}, L2GCN \cite{zhang2019bayesian} or BGCN \cite{zhang2019bayesian}) are initially used to recognize the emotion state, shown in Table \ref{tab:sota_smg}. \hl{Different from the deep models \cite{Yan2018SpatialTG, Peng2019LearningGC, Liu2020DisentanglingAU} in an end-to-end way, these models \cite{zhang2019bayesian, zhang2019bayesian, Chen2019AnalyzeSG,Chen2023SMGAM} recognized MiGs firstly and then the emotion states based on MiGs, which exploits a two-stage learning strategy to obtain the model.} In these methods, using GCN based models could not achieve good performance as GCN is more suitable to extract features, rather than performing the inference. The classifiers like BayesianNet and WSGN would achieve better results as the direct inference has been performed. In one word, these two-stage recognition pipeline could simplify the recognition procedure while the performance would be limited by the recognition of MiGs even the recognition of MiGs is improved greatly recently \cite{Huang2023MicrogestureCB}. For the methods in an end-to-end way which learn the features and classifiers jointly \hl{like \cite{Yan2018SpatialTG, Peng2019LearningGC, Liu2020DisentanglingAU}, it becomes difficult to learn representative features from the whole body joints}. The samples from SMG has larger variation for the full body having more meaningless gestures like shaking legs. So the reconstruction of MiGs is more difficult and the model would focus more on the recognition of emotion states by using different loss weights compared to the iMiGUE dataset. With the enhanced hyperedges, the performance of our proposed H2OFormer on SMG dataset is greatly improved, and it outperforms other methods.

\hl{Reading the emotions from the MiGs is not only difficult to the deep models but also challenging for the human beings. Investigated by \cite{Chen2023SMGAM} on SMG dataset, the ordinary college students and  university staff without any related knowledge were recruited and evaluated as common people, while another three university staff with psychological knowledge were trained to recognize MiGs as human experts. In the existing works, the learning based methods could purse the similar emotion recognition ability of common people, while the experts achieve higher performance with a clear gap than learning based methods. In this work, with using the hyperedge representation, the performance of deep learning based method could reach the level of human experts. From the results of Table \ref{tab:sota_smg}, it can be concluded that it is the first time for the learning based method to achieve the performance comparable to human experts. However, the human experts trend to multiple cues such as facial expressions and even overall impressions from appearance based data (RGB data) to determine the emotional stress states \cite{Chen2023SMGAM}. So the cognition level of learning based methods has been promoted from the common people to the professional experts by our proposed H2OFormer.}

\subsection{Discussion}
\hl{From the comparison and ablation results, it can be observed that our proposed method by leveraging the hyperedge representation can achieve the best performance of $0.70+$ accuracy beyond all existing learning based methods in different scenarios. The existing methods based on GCNs for skeleton data have explored the temporal motion information, multi-scale information and the architecture search strategy to continuously improve the recognition performance, while our work tries a different way for further considering the relationships of body joints. It shows that the combination of the dynamic hyperedge based self-attention and the one-stage learning in an encoder-decoder-like architecture could improve the recognition ability for revealing the hidden emotions. Especially, the usage of effective hyperedge representation provides a new perspective to explore the motion information by considering the relationships of various body joints. Besides, this work implies that the deep learning based technique by incorporating more modules may improve the recognition ability of hidden emotions. With continued progress of recognition accuracy from 0.4+ to 0.7+ in recent years, our vision based method has caught up with the level of humans. This will greatly enhance confidence in the use of the technology with being helpful to humans.

Although our proposed model achieves promising performance for the MiG based emotion recognition, some disadvantages still exist by observing the ablation study experiments. Firstly, the architecture of the deep model is not very robust. With various blocks of encoder and decoder in H2OFormer, the performance still changes with a certain degree of randomness. It has not show the obvious increase or decrease rules with adding or reducing more blocks. It will become tricky to choose a suitable number of blocks for new application scenarios. Secondly, our proposed method promotes the recognition ability by introducing more computation with designing a dynamic hyperedge representation in each block compared to the existing hypergraph based models. It obviously increases requirements for hardware resources and training time.}

\section{Conclusion and Future Work}\label{sec:con}
This paper proposed hybrid-supervised hypergraph-enhanced Transformer for micro-gesture based hidden emotion recognition. In the proposed method, the Transformer based encoder and decoder were separately designed by utilizing the hypergraph-enhanced self-attention for reconstructing the local and subtle MiGs. In the Transformer, the hyperedges between body joints were \hl{literally} updated to enhancing the representation of the relationships of local motion. Based on the motion reconstruction, a recognition head from the encoder was further constructed for modeling the micro-gestures and the emotion states. The results on ablation-study and comparison to SOTAs showed that the proposed method achieved the best performance on two publicly well-known datasets. \hl{In the future work, we will explore a more effective module for measuring the relationship of different body joints and design a robust architecture based on the action generation technique.}

\bibliographystyle{IEEEtran}
\bibliography{main}

\end{document}